\documentclass[11pt]{article}

\usepackage[final]{acl}
 
\usepackage{times}
\usepackage{latexsym}

\usepackage[T1]{fontenc}

\usepackage[utf8]{inputenc}

\usepackage{microtype}

\usepackage{inconsolata}

\usepackage{graphicx}

\usepackage{amsmath}
\usepackage{amssymb}
\usepackage{placeins}
\usepackage{graphicx}
\usepackage{listings}
\usepackage[most]{tcolorbox}
\usepackage{forest}
\usepackage{enumitem}
\usepackage{graphicx}

\tcbset{
tracecasewide/.style={
  enhanced,
  colback=gray!2,
  colframe=gray!40,
  colbacktitle=gray!12,
  coltitle=black,
  boxrule=0.35pt,
  arc=1.2pt,
  left=5pt,
  right=5pt,
  top=4pt,
  bottom=4pt,
  fonttitle=\bfseries\footnotesize
}
}

\forestset{
trace wide tree/.style={
  for tree={
    grow'=0,
    parent anchor=east,
    child anchor=west,
    anchor=west,
    align=left,
    font=\tiny\sffamily,
    inner sep=0.8pt,
    l sep=7pt,
    s sep=1.5pt,
    edge={draw=blue!55!black, line width=0.3pt},
    if n children=0{
      fill=blue!4,
      draw=blue!35,
      rounded corners=1pt,
    }{},
  }
}
}
\usepackage{multirow}
\usepackage{booktabs}
\usepackage[most]{tcolorbox}
\usepackage{makecell}
\usepackage{tabularx}
\usepackage{array}
\usepackage{subcaption}
\newcolumntype{Y}{>{\centering\arraybackslash}X}
\usepackage{url}

\title{
TRACE: Deployable Tree-Relational Structure Enhancement for \\
Oncology LLMs
}
\author{
  \textbf{Jizheng Lai\textsuperscript{1}},
  \textbf{Yingyun Li\textsuperscript{2}},
  \textbf{Ying Qin\textsuperscript{2}},
  \textbf{Haiyang Qian\textsuperscript{1}}
  \thanks{Corresponding author.}
  \\
  \textsuperscript{1}Pharmaron, China
  \\
  \textsuperscript{2}AI Starfish, China
  \\
  \texttt{\{jizheng.lai,haiyang.qian\}@pharmaron.com}
  \\
  \texttt{\{yingyun.li,ying.qin\}@aistarfish.com}
}
\begin{document}
\maketitle

\begin{abstract}
Large language models are increasingly used in oncology applications, but their predictions are often weakly grounded in explicit medical structure. We present TRACE, a deployable tree-relational enhancement framework for oncology LLMs. TRACE separates expensive offline structure learning from lightweight online inference: oncology concepts and relations are organized into an updatable tree-relational structure, refined using LM-loss-derived evidence, and retrieved at inference time as compact prompt evidence. This design supports task-adaptive evidence selection without requiring supervised labels in the zero-shot setting. Across ten oncology classification tasks and one MedQuAD CancerGov QA benchmark, TRACE improves both label-free evaluation and supervised fine-tuning. Additional analyses show that TRACE improves over vanilla RAG and generic GraphRAG, remains useful under leakage-controlled METABRIC inputs, and produces interpretable evidence paths aligned with clinical reasoning. These results suggest that explicit, updatable medical structure is a practical path toward more accurate and auditable oncology LLM deployment.
\end{abstract}

\section{Introduction}

Large language models (LLMs) are increasingly used as interfaces for medical knowledge and clinical decision support. Recent medical LLMs and evaluations show rapid progress in medical question answering and consultation-style reasoning, but also reveal persistent gaps between benchmark performance and robust real-world clinical reasoning~\cite{singhal2023medpalm,zhang2023huatuogpt,liu2024clinicbench,wang2025baichuanm1,m2team2025baichuanm2,m3team2026baichuanm3,ren2026medicalreasoning}. These observations suggest that medical LLMs should not only store domain knowledge parametrically, but also identify and prioritize the clinical evidence needed for a specific decision.

This requirement is particularly important in oncology. Cancer reasoning often depends on structured interactions among disease categories, histology, molecular markers, staging criteria, treatment, recurrence, and prognosis. Curated oncology resources such as OncoKB and CIViC organize genomic alterations with diagnostic, prognostic, and therapeutic implications~\cite{chakravarty2017oncokb,griffith2017civic}, while biomedical knowledge graphs such as PrimeKG integrate heterogeneous biomedical entities and relations for precision-medicine analysis~\cite{chandak2023primekg}. These resources highlight the value of explicit medical structure for oncology reasoning.

Retrieval-augmented generation (RAG) grounds LLMs in external evidence, and recent graph-based RAG methods retrieve triples, paths, or subgraphs for multi-hop reasoning and hallucination mitigation~\cite{xiong2024medrag,he2024gretriever,gutierrez2024hipporag,li2025subgraphrag,wu2025medgraphrag}. However, most existing systems treat the external structure as a fixed retrieval index. They retrieve passages or graph fragments at inference time, but do not explicitly refine the medical structure using model-side evidence. Recent clinical knowledge-graph construction studies further suggest that medical graphs should be continuously updated as evidence changes~\cite{das2026clinicalkg,xu2026openworldkgrag}.

We propose TRACE, a Tree-Relational Adaptive Concept Engine for Oncology LLMs. TRACE is designed as a deployable structure-enhancement framework rather than a new monolithic model. The expensive stage, including concept-query retrieval, structure-oriented adaptation, and LM-loss-based structure refinement, is performed offline. At inference time, TRACE only retrieves compact tree paths and graph relations and injects them into the prompt, making the online path lightweight and compatible with existing LLM serving stacks.

TRACE represents oncology knowledge as a hierarchical concept tree and a relation graph over the same concept set. The tree captures coarse-to-fine medical organization, while the graph captures cross-branch clinical relations. The structure is refined using language-model evidence and then reused for label-free zero-shot inference, supervised fine-tuning, and deployment-oriented evidence-budget control.

Our contributions are:
\begin{itemize}
    \item We introduce TRACE as a deployable tree-relational enhancement framework for oncology LLMs, separating offline structure learning from lightweight online retrieval and prompt augmentation.
    \item We refine tree nodes and graph edges with LM-loss-derived Bayesian updates, and show through initial-vs-refined structure ablation that this refinement improves knowledge-intensive oncology QA.
    \item We evaluate TRACE against strong LLMs, vanilla RAG, and generic GraphRAG under the same oncology corpus, showing that the learned tree-relational structure provides gains beyond adding retrieved text.
    \item We evaluate deployment-relevant properties of TRACE, including leakage-controlled robustness, inference overhead, and offline structure-update cost, showing that TRACE can be integrated as a lightweight online enhancement with offline maintenance.
\end{itemize}

\section{Method}

\begin{figure*}[!h]
    \centering
    \includegraphics[width=\textwidth]{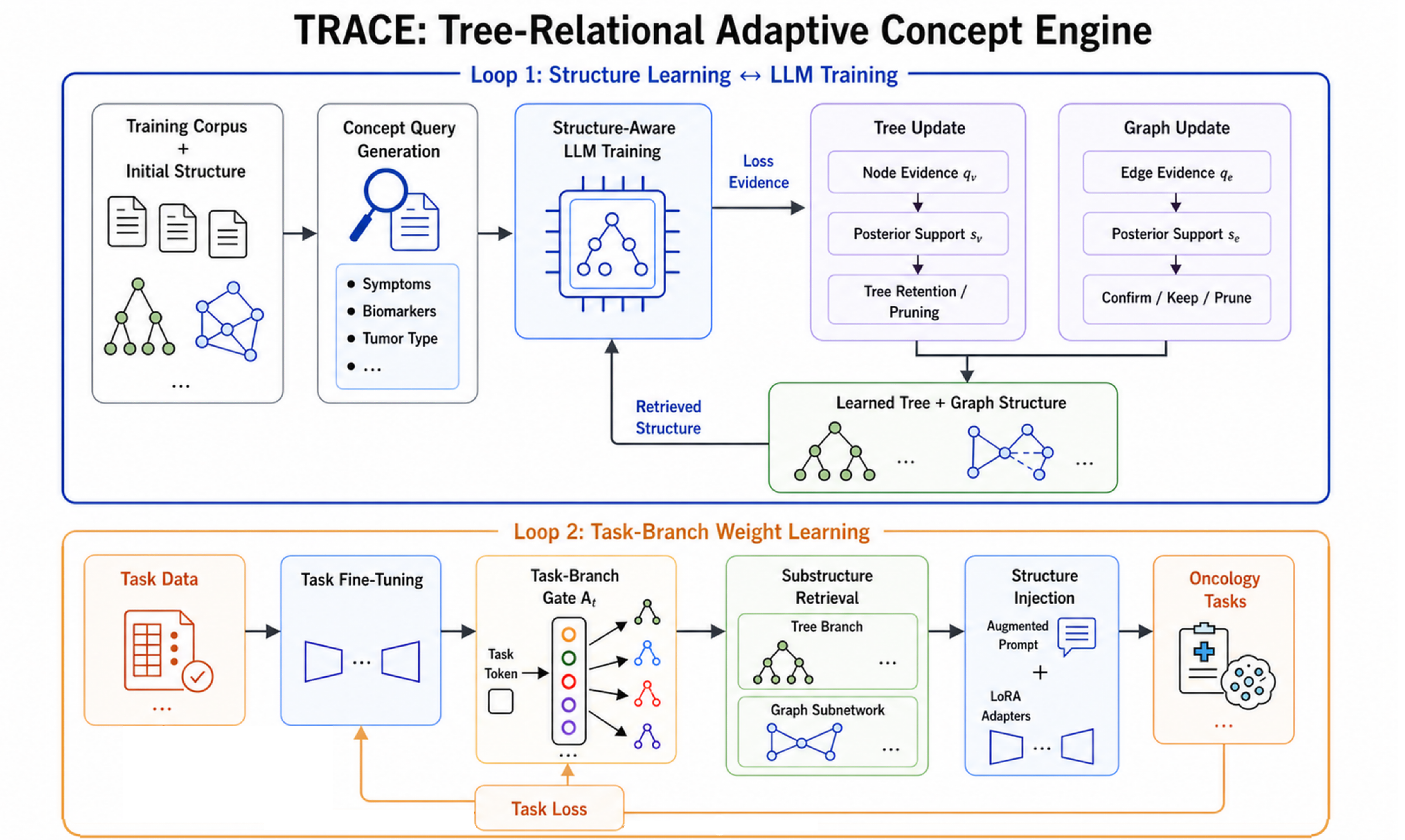}
    \caption{
    Overview of TRACE. Offline, TRACE retrieves concept-query corpus samples, performs structure-oriented LoRA adaptation, estimates LM-loss evidence, and updates tree nodes and graph edges with Bayesian rules. Online, TRACE retrieves compact tree-relational evidence and injects it into downstream prompts. Supervised settings can additionally learn task-branch weights.
    }
    \label{fig:workflow}
\end{figure*}

TRACE consists of offline structure learning and online task integration. Unlike retrieval-only GraphRAG, which reads a fixed external graph at inference time, TRACE treats the structure as an offline learnable artifact. Structure-oriented adaptation produces LM-loss evidence, which is written back as node and edge support through a Beta-Bernoulli update; the refined structure then governs subsequent retrieval. Online inference only retrieves relevant tree paths and graph relations for prompt augmentation, without updating the structure on the serving path. This separation allows the structure to be versioned and maintained independently of the serving backbone. Figure~\ref{fig:workflow} summarizes the system.

\subsection{Tree-Relational Oncology Structure}

Let the oncology concept tree be \(\mathcal{T}=(\mathcal{V}_T,\mathcal{E}_T)\), where nodes are oncology concepts and edges are parent-child relations. Each node \(v\) is associated with a textual representation \(x_v\), including its concept name, aliases, English translation, node type, and full tree path. Inspired by the thematic grouping of the minimal Common Oncology Data Elements (mCODE) standard~\citep{mcode2025}, the tree covers major oncology branches such as cancer type, symptoms, pathology, staging, biomarkers, treatment, and outcome.

On top of the tree, TRACE defines a relation graph \(\mathcal{G}=(\mathcal{V}_G,\mathcal{E}_G)\), where \(\mathcal{V}_G\) shares the same concept inventory. Each edge is written as \(e=(u,r,v,w_e)\), where \(u\) and \(v\) are concepts, \(r\) is a clinical relation type, and \(w_e\in[0,1]\) is the edge weight. The initial structure is built from oncology corpora and label-free task-facing input text; answer-like fields and target labels are removed whenever task data are used for structure construction.

To obtain model-side evidence, TRACE uses each node and edge as a concept query to retrieve relevant oncology corpus passages. The retrieved passages are converted into structure-oriented training samples and used to train a lightweight low-rank adaptation (LoRA) adapter, which we refer to as Concept Query LoRA. After this adaptation, canonical node and edge statements are scored by causal language modeling loss. The loss is converted into a relative soft evidence score:
\[
q_z=\sigma\left(\frac{\ell_{\mathrm{med}}-\ell_z}{\tau}\right),
\]
where \(z\) denotes a node or edge, \(\ell_z\) is its LM loss, \(\ell_{\mathrm{med}}\) is the median loss over elements of the same type, \(\sigma(\cdot)\) is the sigmoid function, and \(\tau\) is a temperature. A higher \(q_z\) means the structural statement is better supported by the adapted model relative to other statements of the same type.

\subsection{Bayesian Structure Refinement}

TRACE keeps the tree topology fixed and estimates a posterior support score for each node. For node \(v\), evidence is aggregated from the node and its descendants:
\[
\tilde{q}_v=\max\bigl(\{q_v\}\cup\{q_u:u\in\mathrm{Desc}(v)\}\bigr),
\]
where \(\mathrm{Desc}(v)\) denotes the set of descendants of node \(v\). This allows a strongly supported fine-grained concept to also support its ancestors, which define the tree path needed to interpret it.

To update node support using a Beta-Bernoulli Bayesian model, we model node usefulness as a latent Bernoulli variable with a Beta prior distribution $\mathrm{Beta}(\alpha_v,\beta_v)$. Given prior support \(p_T\), prior strength \(c_T\), and smoothing coefficient \(\rho\), the prior pseudo-counts are
\[
\alpha_v=1+c_Tp_T,\qquad
\beta_v=1+c_T(1-p_T).
\]
The Beta distribution is conjugate to the Bernoulli likelihood, so posterior inference can be performed by updating the two pseudo-counts of supporting and opposing evidence. 
The LM-derived score \(\tilde{q}_v\in[0,1]\) is treated as fractional positive evidence and \(1-\tilde{q}_v\) as fractional negative evidence:
\[
\alpha'_v=\rho\alpha_v+\tilde{q}_v,\qquad
\beta'_v=\rho\beta_v+(1-\tilde{q}_v).
\]
The posterior node support is the posterior mean:
\[
s_v=\frac{\alpha'_v}{\alpha'_v+\beta'_v}.
\]
Nodes with high support are prioritized during retrieval, while low-support nodes remain as background ontology.

Graph edges are updated analogously, but unlike tree nodes, each graph edge already has an initial confidence weight. For edge \(e=(u,r,v,w_e)\), where \(u\) and \(v\) are concepts, \(r\) is the relation type, and \(w_e\) is the initial edge weight, TRACE initializes a Beta prior from \(w_e\), combines it with LM-derived edge evidence \(q_e\), and obtains the posterior edge support:
\[
s_e=\frac{\alpha'_e}{\alpha'_e+\beta'_e}.
\]
We use this posterior support as the updated edge weight, i.e., \(w'_e=s_e\), for downstream retrieval. Edges with high posterior support are confirmed, uncertain edges are retained as candidates, and weak edges are excluded from downstream retrieval. Detailed update equations and thresholds are provided in Appendix~\ref{app:implementation}. 

We use Beta-Bernoulli refinement as a bounded and interpretable smoothing mechanism, rather than assuming that it is intrinsically more accurate than deterministic weighting. It represents initial structural confidence as prior pseudo-counts and LM-derived evidence as fractional observations, yielding a closed-form posterior support in \([0,1]\). Appendix~\ref{app:additional-ablation} compares this update with unrefined initial weights and simple linear weighting.

\subsection{Tree-Relational RAG and Task Integration}

Tree-Relational RAG is the online TRACE component. Given a sample \(x_i\), it retrieves concepts, tree paths, and graph edges using only input-side information such as the question, options, patient description, or serialized clinical fields. Retrieval combines sample matching with updated structural confidence:
\[
\mathrm{Score}(x_i,e)=\mathrm{Match}(x_i,e)\cdot w'_e,
\]
where \(\mathrm{Match}(x_i,e)\) captures lexical overlap, keyword hits, and endpoint matches. The selected evidence is appended as auxiliary context. Classification uses candidate log-likelihood scoring, while generation uses standard decoding.

In supervised fine-tuning, TRACE can also learn task-branch weights:
\[
\pi_t=\mathrm{softmax}(A_t),\qquad
\]
\[
\mathrm{Score}_t(x_i,e)=\mathrm{Score}(x_i,e)\cdot \mathrm{BranchRel}_t(e).
\]
The branch logits are warm-started from label-free evidence and optimized with LoRA parameters. Zero-shot TRACE does not use learned task-branch weights.

\section{Experiments}

\paragraph{Datasets and setup.}

We evaluate TRACE on ten oncology classification tasks and one QA generation benchmark: three Chinese oncology examination QA datasets, two private de-identified clinical-record tasks, five METABRIC-derived breast cancer tasks from cBioPortal \texttt{brca\_metabric}~\citep{curtis2012metabric,pereira2016metabric}, and MedQuAD CancerGov QA~\citep{benabacha2019medquad}. The oncology structure is built from textbooks, clinical guidelines, expert consensus documents, and biomedical literature. 

Unless otherwise specified, TRACE uses Qwen3.5-27B as the backbone, with tree-relational evidence injected through task-adaptive retrieval and prompt augmentation. Beyond accuracy, we evaluate robustness to noisy EMR-style inputs, inference latency, and structural evidence traces, since deployment requires stable behavior under imperfect clinical text and bounded serving cost. Dataset statistics, preprocessing, split construction, leakage control, metrics, and hyperparameters are provided in Appendices~\ref{app:datasets}, \ref{app:metrics}, and~\ref{app:implementation}.

\subsection{Zero-Shot Oncology Task Evaluation}

\begin{table*}[t]
\centering
\footnotesize
\setlength{\tabcolsep}{3.2pt}
\renewcommand{\arraystretch}{1.08}
\begin{tabular*}{\textwidth}{@{\extracolsep{\fill}}>{\raggedright\arraybackslash}p{2.25cm}cccccccc@{}}
\toprule
Task
& \makecell{Baichuan-\\M2-32B}
& \makecell{Baichuan-\\M3-\\235B}
& \makecell{Llama-4-\\Scout}
& \makecell{HuatuoGPT-\\3-32B}
& \makecell{Qwen3-\\32B}
& \makecell{DeepSeek-\\V4-\\Flash}
& \makecell{Qwen3.5-\\27B}
& TRACE \\
\midrule
\multicolumn{9}{l}{\textit{Oncology examination QA}} \\
Professional Practice & 0.785 & \underline{0.840} & 0.774 & 0.796 & 0.807 & \underline{0.840} & 0.818 & \textbf{0.906} \\
Oncology Attending & 0.727 & 0.787 & 0.628 & 0.740 & 0.749 & \textbf{0.815} & 0.781 & \underline{0.807} \\
Related Knowledge & 0.792 & 0.833 & 0.657 & 0.790 & 0.794 & \textbf{0.853} & 0.836 & \underline{0.845} \\
\midrule
\multicolumn{9}{l}{\textit{Private clinical records}} \\
Common Cancer Dx. & 0.701 & \textbf{0.733} & 0.649 & 0.710 & 0.718 & \underline{0.725} & 0.718 & \underline{0.725} \\
Breast Stage Pred. & 0.453 & 0.502 & 0.273 & 0.498 & 0.502 & \underline{0.505} & 0.491 & \textbf{0.519} \\
\midrule
\multicolumn{9}{l}{\textit{METABRIC public benchmark}} \\
Stage & 0.604 & 0.602 & 0.569 & 0.532 & 0.556 & 0.637 & \textbf{0.729} & \underline{0.718} \\
Histology & \underline{0.762} & \underline{0.762} & 0.761 & \textbf{0.763} & \underline{0.762} & \underline{0.762} & \underline{0.762} & \underline{0.762} \\
ER Status & 0.723 & 0.753 & 0.739 & 0.739 & 0.768 & 0.714 & \underline{0.804} & \textbf{0.823} \\
PR Status & \underline{0.744} & 0.743 & 0.534 & 0.741 & 0.733 & 0.696 & \textbf{0.745} & \underline{0.744} \\
Relapse Status & 0.624 & 0.624 & 0.599 & 0.564 & 0.548 & \textbf{0.631} & 0.601 & \underline{0.627} \\
\midrule
\multicolumn{9}{l}{\textit{Group-level macro averages}} \\
Exam QA Avg. & 0.768 & 0.820 & 0.686 & 0.775 & 0.783 & \underline{0.836} & 0.811 & \textbf{0.853} \\
Private Clinical Records Avg. & 0.582 & \underline{0.617} & 0.461 & 0.604 & 0.610 & 0.615 & 0.604 & \textbf{0.622} \\
METABRIC Avg. & 0.692 & 0.697 & 0.640 & 0.668 & 0.673 & 0.688 & \underline{0.728} & \textbf{0.735} \\
\midrule
Task-macro Avg.
& 0.693 & 0.718 & 0.618 & 0.687 & 0.694 & 0.718 & \underline{0.728} & \textbf{0.748} \\
\bottomrule
\end{tabular*}
\caption{
Label-free zero-shot accuracy on ten oncology classification tasks.
TRACE uses Qwen3.5-27B as the backbone and injects retrieved tree-relational evidence without supervised training on target-task labels.
All baselines are evaluated under the same label-free setting.
Each dataset block reports per-task accuracy and a group-level macro average, and the last row reports the macro average over all ten tasks.
Bold and underline denote the best and second-best results in each row.
}
\label{tab:zeroshot-main}
\end{table*}

We use ``zero-shot'' to denote the absence of supervised training on target-task labels. This includes plain LLM inference as well as label-free structure-adapted variants: structure learning uses oncology corpora and input-side text only, while retrieved evidence is constructed without task answers.

Table~\ref{tab:zeroshot-main} reports label-free zero-shot accuracy on ten oncology tasks. We compare TRACE, instantiated with Qwen3.5-27B as its backbone, with representative open-weight and API-based LLMs. For closed-set tasks, prediction is made by candidate log-likelihood scoring.

Table~\ref{tab:zeroshot-main} shows that TRACE achieves the best task-macro average across all ten tasks, improving over Qwen3.5-27B from 0.728 to 0.748. Gains are observed across oncology examination QA, private clinical records, and METABRIC, suggesting that explicit structural evidence benefits both knowledge-intensive QA and structured clinical-record classification.

The results also show that TRACE is not merely benefiting from a stronger base model. DeepSeek-V4-Flash remains the strongest individual baseline on several examination tasks, while Qwen3.5-27B is strong on METABRIC stage and receptor-status prediction. TRACE nevertheless obtains the best overall average by making relevant concepts and relations explicit at inference time. Group-level task-macro F1 results and bootstrap confidence intervals are reported in Appendix~\ref{app:zeroshot-f1} and Appendix~\ref{app:main-ci}. TRACE has the highest point estimate across all dataset groups and the 10-task macro average, with the clearest separation on examination QA and the overall macro score. Per-task gains over Qwen3.5-27B are reported in Appendix~\ref{app:full-results}. 

\subsection{Fine-Tuning on Public Oncology Benchmarks}

We next evaluate whether TRACE remains effective when task supervision is available. We use two public benchmarks: METABRIC-derived breast cancer classification and MedQuAD CancerGov question answering. For METABRIC, the five task training splits are concatenated and shuffled, and each model is fine-tuned on the mixed training set before task-wise evaluation. For CancerGov QA, models are fine-tuned on the public document-level split and evaluated by greedy generation.

\begin{table}[t]
\centering
\footnotesize
\setlength{\tabcolsep}{3.2pt}
\renewcommand{\arraystretch}{1.06}
\begin{tabularx}{\columnwidth}{@{}Xcccc@{}}
\toprule
Model / Method
& \makecell{Overall\\Acc.}
& \makecell{Overall\\F1}
& \makecell{Task-macro\\Acc.}
& \makecell{Task-macro\\F1} \\
\midrule
Qwen3-32B LoRA & 0.755 & 0.389 & 0.758 & 0.546 \\
HuatuoGPT-3-32B LoRA & 0.748 & 0.364 & 0.751 & 0.526 \\
Baichuan-M2-32B LoRA & 0.744 & 0.359 & 0.744 & 0.526 \\
Qwen3.5-27B LoRA & 0.757 & 0.361 & 0.760 & 0.528 \\
Llama-4-Scout LoRA & 0.758 & 0.360 & 0.761 & 0.528 \\
TRACE & \textbf{0.797} & \textbf{0.480} & \textbf{0.797} & \textbf{0.579} \\
\bottomrule
\end{tabularx}
\caption{
Fine-tuning results on the five METABRIC tasks.
Overall metrics are computed by pooling all test examples, while task-macro metrics average the corresponding metric across the five tasks.
Bold denotes the best result in each column.
}
\label{tab:metabric-ft}
\end{table}

Table~\ref{tab:metabric-ft} shows that TRACE improves METABRIC task-macro accuracy from the strongest baseline value of 0.761 to 0.797, and task-macro F1 from 0.546 to 0.579. On CancerGov QA, TRACE also obtains the best ROUGE-L and BERTScore; full results are reported in Appendix~\ref{app:cancerqa-ft}. These results indicate that task supervision further benefits from explicit tree-relational evidence.

\subsection{Ablation Study}

We summarize the main ablations in Table~\ref{tab:main-ablation-summary}. The first block decomposes TRACE into Concept Query LoRA and Tree-Relational RAG. Concept Query LoRA alone gives limited direct task gains because its primary role is to support structure learning, whereas Tree-Relational RAG provides the main zero-shot improvement by injecting retrieved structural evidence. The second block compares TRACE with retrieval baselines built from the same oncology corpus. TRACE outperforms both vanilla passage RAG and generic GraphRAG on the 10-task macro average, showing that the gain is not simply due to adding external context. The third block isolates LM-loss Bayesian refinement by comparing TRACE with the initial unrefined structure. Refinement mainly improves examination QA, suggesting that updated structural reliability is most useful for knowledge-intensive reasoning tasks.

\begin{table}[t]
\centering
\footnotesize
\setlength{\tabcolsep}{3pt}
\renewcommand{\arraystretch}{1.04}
\begin{tabularx}{\columnwidth}{@{}Xcccc@{}}
\toprule
Method
& \makecell{Exam\\QA}
& \makecell{Clinical\\records}
& METABRIC
& \makecell{10-task\\Avg.} \\
\midrule
Qwen3.5-27B & 0.811 & 0.604 & 0.728 & 0.728 \\
+ Concept Query LoRA & 0.816 & 0.610 & 0.714 & 0.723 \\
+ Tree-Relational RAG & 0.847 & 0.615 & \textbf{0.736} & 0.745 \\
TRACE & \textbf{0.853} & \textbf{0.622} & 0.735 & \textbf{0.748} \\
\midrule
Vanilla RAG & 0.817 & 0.623 & 0.727 & 0.733 \\
Generic GraphRAG & 0.828 & \textbf{0.624} & 0.725 & 0.736 \\
\midrule
Initial structure & 0.838 & \textbf{0.624} & \textbf{0.738} & 0.745 \\
Refined structure & \textbf{0.853} & 0.622 & 0.735 & \textbf{0.748} \\
\bottomrule
\end{tabularx}
\caption{
Main ablation summary.
All numbers are group-level or 10-task macro accuracy.
Bold denotes the best result within each comparison block.
}
\label{tab:main-ablation-summary}
\end{table}

We further ablate structural views in Appendix~\ref{app:tree-graph-ablation}. Tree-only, graph-only, and tree-plus-graph variants all improve over the base model, with the combined structure achieving the best 10-task macro accuracy.

Because METABRIC is retrospective, non-target fields may include downstream treatment or outcome variables. We therefore construct a stricter METABRIC setting that removes treatment- and outcome-related fields, including endocrine therapy, chemotherapy, radiotherapy, surgery, recurrence, survival, follow-up, and death-related fields. As shown in Appendix~\ref{app:metabric-leakage-control}, TRACE remains best among Qwen3.5-27B, Vanilla RAG, and Generic GraphRAG, improving task-macro accuracy to 0.729 and task-macro F1 to 0.527. This suggests that TRACE does not rely solely on reverse-causal treatment or outcome shortcuts.

\subsection{Deployment-Oriented Evaluation}

TRACE separates offline structure maintenance from online serving. For the 300-concept structure, one complete refinement round takes approximately 97 minutes, or 3.1--3.3 H200 GPU-hours; most of this cost comes from LoRA training rather than Bayesian write-back. The write-back scales as \(O(|\mathcal{V}|+|\mathcal{E}|)\), and a linear cost projection estimates approximately 28 and 110 GPU-hours for structures containing 5,000 and 20,000 elements, respectively, although these larger scales were not empirically profiled. Online structural retrieval has a median latency of 2.71 ms, while prompt augmentation increases end-to-end latency from 305.0 ms to 404.2 ms in direct HF/PEFT inference. Detailed measurements and assumptions are reported in Appendices~\ref{app:implementation} and~\ref{app:deployment-eval}.

\subsection{Structural Evidence Case Study}

Appendix~\ref{app:case-study} provides two representative cases. In a lung-cancer diagnostic workup question, TRACE retrieves tree evidence covering symptoms, smoking history, thoracic cancer, and lung cancer, together with graph relations linking these clues to bronchoscopy. In a METABRIC stage-prediction example, where treatment and outcome fields are removed, TRACE retrieves staging-related tree nodes and graph relations that connect tumor size and lymph-node status to early-stage breast cancer. These cases illustrate the auditability of TRACE: tree paths expose the relevant clinical hierarchy, while graph edges identify the relations that support the final prediction.

\section{Conclusion}

We presented TRACE, a deployable tree-relational enhancement framework for oncology LLMs. TRACE learns and refines an explicit oncology structure offline, then performs lightweight online retrieval and prompt augmentation for downstream prediction. Across zero-shot, fine-tuning, retrieval-baseline, leakage-controlled, and deployment-oriented evaluations, TRACE improves accuracy while preserving interpretable structural evidence. The results suggest that explicit, updatable medical structure can make oncology LLMs more accurate, auditable, and practical for constrained clinical deployment settings.

\section*{Limitations}

TRACE does not assume that a complete medical ontology is available, since its initial structure is constructed from domain corpora. However, the current refinement stage only recalibrates represented nodes and edges and does not itself recover elements absent from the candidate structure. Structural coverage can nevertheless be extended offline by rerunning or incrementally updating candidate extraction as the corpus evolves, consistent with prior incremental concept-graph learning from text~\cite{lai2024conditional}; this expansion remains separate from the current confidence-refinement stage.

TRACE also uses prompt-based evidence injection, whose effect varies across backbones. Some models benefit consistently from retrieved structure, while others show negative transfer on certain tasks, suggesting that evidence formatting and calibration require further study.

Our evaluation is limited to oncology tasks, and we do not claim cross-specialty generalization. Moreover, the retrieved evidence has not been systematically assessed by clinical experts for correctness, relevance, harmful omissions, or safety. Our results therefore establish benchmark performance and evidence inspectability, rather than prospective clinical validity.

\section*{Ethical Considerations}

Private clinical-record tasks are constructed from de-identified records, and task-facing data used for structure construction or retrieval exclude labels, reference answers, outputs, and target fields to reduce answer leakage. Because retrieved evidence and model predictions may still be incomplete, outdated, or incorrect, any real-world use would require institutional approval, privacy and security review, prospective validation, workflow integration studies, and final review by qualified clinicians. TRACE should be viewed as a decision-support system rather than a substitute for clinical judgment.

\bibliography{custom}
\clearpage

\appendix

\section{Implementation Details}
\label{app:implementation}

\paragraph{Main implementation settings.}
Unless otherwise stated, TRACE uses Qwen3.5-27B as the backbone language model. The learned structure used in the main experiments is the updated 300-concept tree-relational structure. Downstream structure injection uses direct prompt evidence without evidence compression or reranking. The default retrieval budget is six concepts and six graph edges per sample. Supervised task LoRA fine-tuning uses one training epoch, learning rate \(2\times10^{-5}\), batch size 1, gradient accumulation 8, maximum sequence length 1536, bfloat16 precision, and LoRA rank 16 with LoRA scaling parameter \(\alpha_{\mathrm{LoRA}}=32\) and dropout 0.05. The LoRA target modules are all linear layers.

\paragraph{LM-loss evidence.}
Each tree node and graph edge is verbalized into a canonical causal language modeling sample. For a structural element \(z\), let \(x_z=(x_{z,1},\ldots,x_{z,n})\) be its token sequence and \(m_j\) be the attention mask. TRACE computes the token-averaged negative log-likelihood:
\[
\ell_z
=
-\frac{1}{\sum_{j=2}^{n}m_j}
\sum_{j=2}^{n}
m_j\log p_\theta(x_{z,j}\mid x_{z,<j}).
\]
Here \(\theta\) denotes the adapted language-model parameters. The loss is converted into a relative evidence score:
\[
q_z=\sigma\left(\frac{\ell_{\mathrm{med}}-\ell_z}{\tau}\right),
\]
where \(\ell_{\mathrm{med}}\) is the median loss over structural elements of the same type, \(\sigma\) is the sigmoid function, and \(\tau\) is a temperature parameter. This score is used as soft model evidence rather than as a calibrated factuality probability.

\paragraph{Graph update equations.}
For graph edge \(e=(u,r,v,w_e)\), TRACE initializes a Beta prior from the original edge weight:
\[
\alpha_e=1+c_Gw_e,\qquad
\beta_e=1+c_G(1-w_e),
\]
where \(c_G\) controls the prior strength. Given LM-derived evidence \(q_e\), the posterior pseudo-counts are
\[
\alpha'_e=\rho\alpha_e+q_e,\qquad
\beta'_e=\rho\beta_e+(1-q_e),
\]
where \(\rho\) is the smoothing coefficient shared with the tree update. The updated edge weight is the posterior mean:
\[
w'_e=\frac{\alpha'_e}{\alpha'_e+\beta'_e}.
\]
Edges are then assigned a retrieval status:
\begingroup
\small
\[
\mathrm{status}(e)=
\begin{cases}
\texttt{confirmed}, & w'_e\ge \gamma_{\mathrm{confirm}},\\
\texttt{candidate}, & \gamma_{\mathrm{keep}}\le w'_e<\gamma_{\mathrm{confirm}},\\
\texttt{pruned}, & w'_e<\gamma_{\mathrm{keep}}.
\end{cases}
\]
\endgroup
where \(\gamma_{\mathrm{confirm}}\) and \(\gamma_{\mathrm{keep}}\) are confirmation and retention thresholds.

\paragraph{Iterative structure refinement.}
At round \(r\), TRACE starts from the current structure \((\mathcal{T}^{(r-1)},\mathcal{G}^{(r-1)})\), retrieves concept-query corpus samples, trains or continues the structure-oriented LoRA adapter, and rescores canonical node and edge statements. The Bayesian update then produces \((\mathcal{T}^{(r)},\mathcal{G}^{(r)})\), which is used for the next round.

\paragraph{Prompt evidence format.}
For downstream prediction, TRACE appends a concise evidence block to the input. The block contains retrieved tree nodes with their paths and retrieved graph edges with relation type, edge weight, and evidence text:
\[
\texttt{[Relevant nodes]}:\quad
v,\ \mathrm{path}(v)
\]
\[
\begin{aligned}
\texttt{[Relevant edges]}:\quad&
u\ \texttt{--}r\texttt{-->}\ v,\ w'_e,\\
&\mathrm{evidence}(e).
\end{aligned}
\]
The evidence is treated as auxiliary context, while the original task input remains the primary source. In the main experiments, we use the direct evidence format without compressed evidence or reranking.

\paragraph{Task prompt templates.}
For multiple-choice oncology QA, the prompt contains the question, answer options, and the retrieved structural evidence block. The model is instructed to output only the option letter. For closed-set clinical classification tasks, the prompt contains the serialized input fields, candidate labels, and the retrieved evidence block, and the model is instructed to output exactly one label. In both settings, the prompt explicitly treats structural evidence as auxiliary context: when the retrieved evidence conflicts with the original task input, the original input takes priority.

The multiple-choice prompt follows the template:
\begin{quote}
\small
\textbf{Question:} \(\langle\)question stem\(\rangle\)

\textbf{Options:} \(\langle\)candidate options\(\rangle\)

\textbf{Relevant structural evidence:} \(\langle\)retrieved tree paths and graph edges\(\rangle\)

Please answer with only the option letter.
\end{quote}

The closed-set classification prompt follows the template:
\begin{quote}
\small
\textbf{Patient / sample information:} \(\langle\)serialized non-target fields\(\rangle\)

\textbf{Relevant structural evidence:} \(\langle\)retrieved tree paths and graph edges\(\rangle\)

\textbf{Candidate labels:} \(\langle\)label set\(\rangle\)

Please output exactly one candidate label.
\end{quote}

\paragraph{Task fine-tuning configuration.}
Unless otherwise specified, supervised task LoRA fine-tuning uses one training epoch, learning rate \(2\times10^{-5}\), batch size 1, gradient accumulation 8, maximum sequence length 1536, bfloat16 precision, and LoRA rank 16 with LoRA scaling parameter \(\alpha_{\mathrm{LoRA}}=32\) and dropout 0.05. The LoRA target modules are all linear layers. For direct prompt evidence, the default retrieval budget is six concepts and six graph edges per sample.

\paragraph{Deployment cost of structure construction and update.}
We report the deployment cost for the original structure used in the main experiments. The initial structure is a finalized artifact containing 300 oncology concepts and 280 relations. Its metadata records that it was constructed from 3,757 answer-blind task/corpus items and 590 oncology literature items, with answer, label, reference, candidate, and evidence fields excluded to reduce leakage risk. Since the original construction logs were not retained as separately profiled GPU jobs, we report its scale and provenance rather than estimating GPU-hours for initial construction.

Starting from this finalized structure, we profile one LM-loss Bayesian refinement round on H200 GPUs. The refinement consists of retrieval and update-data preparation, LoRA warmup training, LM-loss scoring, and Bayesian tree/edge weight updates. As shown in Table~\ref{tab:structure-update-cost}, one refinement round takes about 97 minutes end-to-end, with an effective compute cost of approximately 3.1--3.3 GPU-hours. Most of the cost comes from LoRA warmup training, while retrieval, scoring, and Bayesian updates are comparatively lightweight. This cost is incurred offline and only when the structure is refreshed.

The Bayesian update itself scales linearly with the number of structural elements and can be parallelized across independent branches. Under the same evidence budget and measured per-element processing rate, a linear cost projection gives approximately 28 GPU-hours for 5,000 structural elements and 110 GPU-hours for 20,000 elements. These values are analytical projections rather than measurements on structures of those sizes. In a hospital deployment, specialty- or department-level branches can be maintained as separate offline jobs, avoiding the need to retrain a single monolithic medical structure after every local change.

\begin{table*}[t]
\centering
\footnotesize
\setlength{\tabcolsep}{4pt}
\renewcommand{\arraystretch}{1.05}
\begin{tabular*}{\textwidth}{@{\extracolsep{\fill}}p{3.2cm}p{3.2cm}p{2.8cm}ccc@{}}
\toprule
Stage & Input & Output & GPUs & Wall-clock & GPU-hours \\
\midrule
Initial structure construction
& Answer-blind task/corpus text and oncology literature
& 300 concepts / 280 relations
& N/A & Not separately logged & N/A \\
Retrieval and update-data preparation
& 300 concepts / 280 relations
& LM-loss training/scoring corpus
& 2 & 3.3 min & 0.11 \\
LoRA warmup training
& LM-loss corpus
& Structure-adapted LoRA adapter
& 2 & 87.8 min & 2.93 \\
LM-loss scoring and Bayesian update
& Initial tree/graph and LoRA adapter
& Updated tree/graph weights
& 1 & 6.1 min & 0.10 \\
\midrule
Total refinement
& 300 concepts / 280 relations
& Refined structure
& -- & 97 min & 3.1--3.3 \\
\bottomrule
\end{tabular*}
\caption{
Deployment cost of structure construction and one LM-loss Bayesian refinement round.
The initial structure is reported by scale and provenance because its original construction logs were not retained as separately profiled GPU jobs.
GPU-hours are computed from measured wall-clock time and the number of GPUs used in each profiled stage.
}
\label{tab:structure-update-cost}
\end{table*}

\paragraph{Inference latency.}
We profile TRACE inference without vLLM using HuggingFace Transformers and PEFT to directly load Qwen3.5-27B with the LoRA adapter on a single GPU. The profiling uses 100 samples from the three examination QA tasks and two clinical-record tasks, with 20 samples per task, and generates one output token for each sample. Structural retrieval itself is lightweight, with a median latency of 2.71 ms and a P95 latency of 7.43 ms. End-to-end single-sample latency increases from 304.99 ms for the base prompt to 404.24 ms with TRACE evidence, corresponding to a median overhead of 101.50 ms. This overhead mainly comes from prompt expansion: TRACE adds a median of 610 prompt tokens, increasing the median prompt length from 147.5 to 795.5 tokens. These results show that TRACE shifts structure learning offline, while inference-time cost is dominated by the additional structural evidence tokens rather than retrieval.
\begin{table}[t]
\centering
\footnotesize
\setlength{\tabcolsep}{4pt}
\renewcommand{\arraystretch}{1.05}
\begin{tabular}{lcccc}
\toprule
Metric & Median & P90 & P95 & Mean \\
\midrule
Retrieval latency (ms) & 2.71 & 5.54 & 7.43 & 3.43 \\
Base latency (ms) & 304.99 & 439.31 & 517.84 & 338.01 \\
TRACE latency (ms) & 404.24 & 619.67 & 669.72 & 447.14 \\
TRACE overhead (ms) & 101.50 & 137.89 & 236.90 & 109.12 \\
Base prompt tokens & 147.5 & 630.4 & 848.3 & 328.9 \\
TRACE prompt tokens & 795.5 & 1230.2 & 1438.6 & 942.7 \\
Extra prompt tokens & 610.0 & 652.0 & 666.0 & 613.8 \\
\bottomrule
\end{tabular}
\caption{
Inference latency of TRACE under direct HuggingFace Transformers + PEFT inference.
Statistics are computed over 100 samples from the three examination QA tasks and two clinical-record tasks, with 20 samples per task, using Qwen3.5-27B with the LoRA adapter on a single GPU and \(\texttt{max\_new\_tokens}=1\).
TRACE latency includes structural retrieval and prompt-augmented model inference.
}
\label{tab:inference-latency}
\end{table}
As shown in Table~\ref{tab:inference-latency}, retrieval adds only 2.71 ms median latency, while most overhead comes from the additional structural evidence tokens.

\section{Dataset Details}
\label{app:datasets}

\paragraph{Structure construction corpus.}
The oncology structure is constructed from a heterogeneous corpus of oncology resources, including clinical oncology textbooks, internal medicine and pathology textbooks, clinical guidelines, expert consensus documents, and biomedical literature. The corpus contains 978 files with a total size of approximately 2.8 GB. The main sources include Chinese oncology textbooks such as \emph{Modern Oncology}, \emph{Precision Oncology}, an oncology graduate textbook, guideline-oriented resources derived from Chinese Society of Clinical Oncology (CSCO) and National Comprehensive Cancer Network (NCCN) guidelines, and their reference collections. When task-facing text is used as auxiliary corpus for structure construction, we remove answer-like fields, including gold labels, references, outputs, and target annotations.

\begin{table*}[t]
\centering
\footnotesize
\setlength{\tabcolsep}{5pt}
\renewcommand{\arraystretch}{1.06}
\begin{tabularx}{\textwidth}{@{}l c X@{}}
\toprule
Dataset & \# examples & Description \\
\midrule
\multicolumn{3}{l}{\textit{Oncology examination QA}} \\
Professional Practice & 181 & Chinese examination-style oncology questions. \\
Oncology Attending Physician & 1,581 & Chinese oncology attending-physician examination questions. \\
Related Professional Knowledge & 708 & Chinese medical professional-knowledge examination questions. \\
\midrule
\multicolumn{3}{l}{\textit{Private de-identified clinical records}} \\
Common Cancer Diagnosis & 131 & De-identified clinical records labeled with major cancer type. \\
Breast Cancer Stage Prediction & 289 & De-identified breast cancer records labeled with stage. \\
\midrule
\multicolumn{3}{l}{\textit{METABRIC public benchmark}}  \\
METABRIC Stage & 1,788 & Stage prediction from non-target METABRIC fields. \\
METABRIC Histology & 2,374 & Histology prediction from non-target METABRIC fields. \\
METABRIC ER Status & 2,469 & ER-status prediction from non-target METABRIC fields. \\
METABRIC PR Status & 1,980 & PR-status prediction from non-target METABRIC fields. \\
METABRIC Relapse Status & 2,488 & Relapse-status prediction from non-target METABRIC fields. \\
\midrule
\multicolumn{3}{l}{\textit{CancerGov QA generation benchmark}} \\
MedQuAD CancerGov & 729 & Cancer question-answer pairs parsed from 116 XML files. \\
\bottomrule
\end{tabularx}
\caption{
Dataset summary. METABRIC tasks are derived from cBioPortal \texttt{brca\_metabric}; CancerGov QA is derived from the MedQuAD CancerGov subset.
}
\label{tab:dataset-summary}
\end{table*}

\paragraph{METABRIC processing.}
We derive five breast cancer classification tasks from METABRIC through cBioPortal \texttt{brca\_metabric}~\citep{curtis2012metabric,pereira2016metabric}. For each task, the target field is withheld from the input, and the remaining clinical, pathological, molecular, treatment, and outcome fields are serialized as the model input. This produces separate tasks for stage, histology, ER status, PR status, and relapse status prediction.

\paragraph{METABRIC input-field control.}
For each METABRIC-derived task, the target field is removed from the serialized input. In the default retrospective setting, all remaining non-target structured fields are retained. In the leakage-controlled setting, we additionally remove treatment-related and outcome-related fields, including endocrine therapy, chemotherapy, radiotherapy, surgery, recurrence, relapse, survival, follow-up time, death status, and other outcome annotations. Details are shown in Table~\ref{tab:metabric-field-control}. This filtered setting is used only for the leakage-control analysis in Table~\ref{tab:metabric-no-treatment-outcome}.
\begin{table}[t]
\centering
\footnotesize
\setlength{\tabcolsep}{4pt}
\renewcommand{\arraystretch}{1.05}
\begin{tabularx}{\columnwidth}{@{}lX@{}}
\toprule
Setting & Input fields \\
\midrule
Default METABRIC
& All non-target structured fields, including clinical, pathological, molecular, treatment, and outcome fields. \\
Leakage-controlled
& Target field removed as before; treatment and outcome fields are additionally removed, including endocrine therapy, chemotherapy, radiotherapy, surgery, recurrence, survival, follow-up, and death-related fields. \\
\bottomrule
\end{tabularx}
\caption{
Input-field control for METABRIC-derived tasks.
}
\label{tab:metabric-field-control}
\end{table}

\paragraph{CancerGov QA processing.}
We use the MedQuAD CancerGov subset~\citep{benabacha2019medquad}. The XML files are parsed into question-answer pairs, and the split is performed at the document level to avoid placing questions from the same source document in both training and test sets. This yields 591 training and 138 test examples.

\paragraph{Leakage control.}
For zero-shot structure construction and evidence retrieval, TRACE uses only input-side information. In classification tasks, label fields and target annotations are excluded from the prompt and from structure construction. In METABRIC tasks, the predicted target field is removed from the serialized input. In CancerGov QA, the answer text is used only as the supervised generation target during fine-tuning, not as retrieval evidence at test time. For supervised fine-tuning experiments, task examples are used only for downstream model training and evaluation; they are not used to construct the concept inventory or relation graph. At inference time, the current sample provides only the retrieval query, while its target field, label, or reference answer remains unavailable.

\paragraph{Data governance and availability.}
The private Chinese clinical-record datasets are de-identified before use and are not publicly released due to institutional and privacy constraints. Apart from these private clinical-record tasks, the training data and task data used in this work can be made publicly available, including the public examination-style oncology QA datasets, METABRIC-derived tasks, MedQuAD CancerGov QA data, and the non-private resources used for structure construction, subject to the license and access policy of each source. TRACE is designed as a research system for oncology language-model evaluation and decision support, not as an autonomous clinical decision-making system.

\section{CancerGov QA Fine-Tuning Results}
\label{app:cancerqa-ft}

Table~\ref{tab:cancerqa-ft} reports supervised fine-tuning results on the MedQuAD CancerGov QA generation benchmark. Models are fine-tuned on the public document-level training split and evaluated with greedy generation on the held-out test split. We report ROUGE-L as a lexical overlap metric and BERTScore as a semantic similarity metric.

\begin{table}[t]
\centering
\footnotesize
\setlength{\tabcolsep}{4pt}
\renewcommand{\arraystretch}{1.06}
\begin{tabularx}{\columnwidth}{@{}Xccc@{}}
\toprule
Model / Method
& ROUGE-L
& \makecell{BERTScore\\Tiny}
& \makecell{BERTScore\\RoBERTa} \\
\midrule
Baichuan-M2-32B LoRA & 0.175 & 0.809 & 0.847 \\
HuatuoGPT-3-32B LoRA & 0.135 & 0.791 & 0.835 \\
Qwen3-32B LoRA & 0.155 & 0.805 & 0.845 \\
Qwen3.5-27B LoRA & 0.254 & 0.828 & 0.872 \\
Llama-4-Scout LoRA & 0.201 & 0.823 & 0.860 \\
TRACE & \textbf{0.278} & \textbf{0.848} & \textbf{0.884} \\
\bottomrule
\end{tabularx}
\caption{
Fine-tuning results on MedQuAD CancerGov QA generation.
Models are evaluated by ROUGE-L and BERTScore.
Bold denotes the best result in each column.
}
\label{tab:cancerqa-ft}
\end{table}

\section{Additional Deployment-Oriented Evaluation}
\label{app:deployment-eval}

We include two additional deployment-oriented evaluations. First, we test robustness under noisy EMR-style inputs, where clinical notes are corrupted with abbreviations, token deletion, word-order perturbation, and irrelevant note fragments while preserving the gold label. Second, we sweep the number of retrieved structural evidence items to quantify the accuracy-latency tradeoff. These experiments are intended to characterize operational behavior rather than replace the main zero-shot and fine-tuning evaluations.

\begin{table*}[t]
\centering
\footnotesize
\setlength{\tabcolsep}{4pt}
\renewcommand{\arraystretch}{1.05}
\begin{tabular*}{\textwidth}{@{\extracolsep{\fill}}llccccc@{}}
\toprule
Noise & Method & Acc. & Macro-F1 & Retention & Tokens & \(\Delta\) Latency (ms) \\
\midrule
Clean & LoRA-only & 0.6779 & 0.5033 & 1.0000 & 392.6 & 0.0 \\
Clean & Tree-Relational RAG & 0.6840 & 0.5030 & 1.0000 & 644.3 & +590.6 \\
Clean & TRACE & \textbf{0.6918} & \textbf{0.5063} & 1.0000 & 990.2 & +1420.9 \\
\midrule
Moderate & LoRA-only & 0.6686 & 0.4914 & 0.9899 & 390.3 & 0.0 \\
Moderate & Tree-Relational RAG & 0.6775 & 0.4939 & 0.9926 & 643.5 & +595.6 \\
Moderate & TRACE & \textbf{0.6921} & \textbf{0.5101} & \textbf{1.0022} & 990.2 & +1425.7 \\
\midrule
Severe & LoRA-only & 0.6386 & 0.4648 & 0.9437 & 385.4 & 0.0 \\
Severe & Tree-Relational RAG & 0.6575 & 0.4714 & \textbf{0.9614} & 638.9 & +595.7 \\
Severe & TRACE & \textbf{0.6618} & \textbf{0.4752} & 0.9568 & 986.6 & +1425.5 \\
\bottomrule
\end{tabular*}
\caption{
Robustness under EMR-like input noise.
Retention is computed relative to each method's clean accuracy.
Latency overhead is measured relative to LoRA-only inference under the same noise condition.
}
\label{tab:dirty-emr}
\end{table*}

\begin{table}[t]
\centering
\footnotesize
\setlength{\tabcolsep}{5pt}
\renewcommand{\arraystretch}{1.05}
\begin{tabular}{ccccc}
\toprule
Top-\(k\) & Acc. & Macro-F1 & Tokens & Latency (ms) \\
\midrule
0  & 0.7213 & 0.5296 & 186.6 & 698.2 \\
2  & 0.7282 & 0.5278 & 355.7 & 1184.1 \\
4  & 0.7313 & 0.5317 & 494.7 & 1592.3 \\
8  & 0.7371 & 0.5330 & 709.7 & 2231.2 \\
12 & 0.7414 & 0.5393 & 819.2 & 2547.3 \\
16 & \textbf{0.7449} & \textbf{0.5408} & 871.1 & 2701.7 \\
\bottomrule
\end{tabular}
\caption{
Accuracy-latency tradeoff as the number of retrieved structural evidence items increases.
Larger evidence budgets improve accuracy but increase prompt length and inference latency.
}
\label{tab:topk-tradeoff}
\end{table}

Table~\ref{tab:dirty-emr} shows that structural evidence improves robustness under noisy clinical inputs. Under severe corruption, LoRA-only drops to 0.6386 accuracy, while Tree-Relational RAG and TRACE reach 0.6575 and 0.6618. TRACE obtains the best accuracy and macro-F1 under all three noise levels, although Tree-Relational RAG provides a cheaper robustness gain with fewer prompt tokens.

Table~\ref{tab:topk-tradeoff} shows that the evidence budget controls the accuracy-latency tradeoff. Accuracy improves monotonically from 0.7213 at \(k=0\) to 0.7449 at \(k=16\), while latency also increases. The improvement begins to saturate after \(k=8\), suggesting that moderate evidence budgets can recover most of the benefit with lower deployment cost.
Latency numbers in this section are measured within each deployment-oriented experiment and should not be directly compared with the single-token latency microbenchmark, because the sampled inputs, prompt lengths, and evaluation settings differ.

\section{Evaluation Protocol and Metrics}
\label{app:metrics}

For closed-set classification tasks, each candidate label is scored by conditional log-likelihood under the model, and the prediction is the candidate with the highest score. We use token-averaged log-likelihood when candidate labels contain multiple tokens:
\[
\hat{y}
=
\arg\max_{y\in\mathcal{Y}}
\frac{1}{|y|}
\sum_{j=1}^{|y|}
\log p_\theta(y_j \mid x, y_{<j}),
\]
where \(x\) is the task input and \(\mathcal{Y}\) is the candidate label set. Accuracy is computed separately for each task. Task-macro accuracy averages task accuracies without weighting by dataset size, whereas overall accuracy pools all test examples before computing accuracy. Macro-F1 is computed within each task, and task-macro F1 averages macro-F1 across tasks. For CancerGov QA generation, we report ROUGE-L and BERTScore as complementary lexical and semantic generation metrics.

\section{Additional Zero-Shot F1 Results}
\label{app:zeroshot-f1}

Table~\ref{tab:zeroshot-group-f1} reports the group-level task-macro F1 corresponding to the zero-shot accuracy results in Table~\ref{tab:zeroshot-main}. 

\begin{table}[h]
\centering
\footnotesize
\setlength{\tabcolsep}{4pt}
\renewcommand{\arraystretch}{1.06}
\begin{tabularx}{\columnwidth}{@{}Xcccc@{}}
\toprule
Model / Method & \makecell{Exam\\QA} & \makecell{Clinical\\records} & METABRIC & \makecell{10-task\\Macro Avg.} \\
\midrule
Baichuan-M2-32B & 0.725 & 0.370 & 0.463 & 0.523 \\
Baichuan-M3-235B & 0.774 & 0.396 & 0.453 & 0.538 \\
Llama-4-Scout & 0.677 & 0.331 & 0.326 & 0.432 \\
HuatuoGPT-3-32B & 0.731 & 0.463 & 0.441 & 0.532 \\
Qwen3-32B & 0.739 & 0.469 & 0.452 & 0.542 \\
DeepSeek-V4-Flash & 0.790 & 0.474 & 0.480 & 0.572 \\
Qwen3.5-27B & 0.765 & 0.460 & \textbf{0.513} & 0.578 \\
TRACE & \textbf{0.803} & \textbf{0.480} & 0.506 & \textbf{0.590} \\
\bottomrule
\end{tabularx}
\caption{
Group-level task-macro F1 for label-free zero-shot evaluation.
Each group column averages macro-F1 over tasks in the corresponding dataset group, and the last column averages macro-F1 over all ten tasks.
Bold denotes the best result in each column.
}
\label{tab:zeroshot-group-f1}
\end{table}

\section{Bootstrap Confidence Intervals for Main Results}
\label{app:main-ci}

We additionally report bootstrap confidence intervals in Table~\ref{tab:main-bootstrap-ci} for the zero-shot main results.
For each method, we perform task-stratified bootstrap resampling: examples are resampled within each task, task-level accuracies are recomputed, and group-level and overall task-macro averages are then computed from the resampled task accuracies.
We use 10,000 bootstrap samples and report 95\% percentile confidence intervals.
This procedure matches the task-macro evaluation protocol used in Table~\ref{tab:zeroshot-main}.

\begin{table*}[t]
\centering
\footnotesize
\setlength{\tabcolsep}{3.6pt}
\renewcommand{\arraystretch}{1.08}
\begin{tabular*}{\textwidth}{@{\extracolsep{\fill}}lcccc@{}}
\toprule
Method
& \makecell{Exam QA\\Acc. (95\% CI)}
& \makecell{Clinical records\\Acc. (95\% CI)}
& \makecell{METABRIC\\Acc. (95\% CI)}
& \makecell{10-task Macro\\Acc. (95\% CI)} \\
\midrule
Baichuan-M2-32B
& 0.768 [0.744, 0.791]
& 0.582 [0.533, 0.630]
& 0.692 [0.683, 0.700]
& 0.693 [0.680, 0.705] \\
Baichuan-M3-235B
& 0.820 [0.798, 0.840]
& 0.617 [0.568, 0.664]
& 0.697 [0.688, 0.706]
& 0.718 [0.706, 0.730] \\
Llama-4-Scout
& 0.686 [0.661, 0.710]
& 0.461 [0.413, 0.509]
& 0.640 [0.631, 0.649]
& 0.618 [0.605, 0.631] \\
HuatuoGPT-3-32B
& 0.775 [0.751, 0.798]
& 0.604 [0.555, 0.652]
& 0.668 [0.659, 0.676]
& 0.687 [0.674, 0.700] \\
Qwen3-32B
& 0.783 [0.760, 0.805]
& 0.610 [0.561, 0.658]
& 0.673 [0.665, 0.682]
& 0.694 [0.681, 0.706] \\
DeepSeek-V4-Flash
& 0.836 [0.815, 0.856]
& 0.615 [0.567, 0.663]
& 0.688 [0.679, 0.697]
& 0.718 [0.706, 0.730] \\
Qwen3.5-27B
& 0.811 [0.789, 0.833]
& 0.604 [0.555, 0.652]
& 0.728 [0.720, 0.736]
& 0.728 [0.716, 0.741] \\
TRACE
& \textbf{0.853 [0.834, 0.870]}
& \textbf{0.622 [0.574, 0.670]}
& \textbf{0.735 [0.727, 0.743]}
& \textbf{0.748 [0.736, 0.759]} \\
\bottomrule
\end{tabular*}
\caption{
Bootstrap confidence intervals for the main zero-shot accuracy results.
Intervals are computed with 10,000 task-stratified bootstrap samples.
Each group score is the macro average over tasks in that group, and the final column is the macro average over all ten tasks.
Bold denotes the best point estimate in each column.
}
\label{tab:main-bootstrap-ci}
\end{table*}

\section{Full Zero-Shot Analysis}
\label{app:full-results}

Table~\ref{tab:trace-qwen35-gain} reports the per-task gain of TRACE over Qwen3.5-27B. TRACE improves most tasks, with the largest gains on Professional Practice, Breast Stage Prediction, relapse status, and ER status. Negative or near-zero changes occur on METABRIC stage and PR status, where Qwen3.5-27B is already a strong baseline.

\begin{table}[t]
\centering
\footnotesize
\setlength{\tabcolsep}{3.2pt}
\renewcommand{\arraystretch}{1.04}
\begin{tabular}{p{2.15cm}ccc}
\toprule
Task & \makecell{Qwen3.5\\-27B} & TRACE & \(\Delta\) \\
\midrule
Professional Practice & 0.818 & 0.906 & +0.088 \\
Oncology Attending & 0.781 & 0.807 & +0.026 \\
Related Knowledge & 0.836 & 0.845 & +0.009 \\
Common Cancer Dx. & 0.718 & 0.725 & +0.007 \\
Breast Stage Pred. & 0.491 & 0.519 & +0.028 \\
METABRIC Stage & 0.729 & 0.718 & -0.011 \\
METABRIC Histology & 0.762 & 0.762 & 0.000 \\
METABRIC ER Status & 0.804 & 0.823 & +0.019 \\
METABRIC PR Status & 0.745 & 0.744 & -0.001 \\
METABRIC Relapse Status & 0.601 & 0.627 & +0.026 \\
\midrule
Task-macro Avg. & 0.728 & 0.748 & +0.020 \\
\bottomrule
\end{tabular}
\caption{
Per-task zero-shot accuracy gain of TRACE over Qwen3.5-27B.
}
\label{tab:trace-qwen35-gain}
\end{table}

\section{Tree--Graph Ablation}
\label{app:tree-graph-ablation}
\begin{table*}[t]
\centering
\footnotesize
\setlength{\tabcolsep}{4.5pt}
\renewcommand{\arraystretch}{1.04}
\begin{tabular*}{\textwidth}{@{\extracolsep{\fill}}lcccccc@{}}
\toprule
Evidence
& \makecell{Prof.\\Practice}
& \makecell{Oncology\\Attending}
& \makecell{Related\\Knowledge}
& \makecell{Common\\Cancer Dx.}
& \makecell{Breast\\Stage Pred.}
& \makecell{5-task\\Macro Avg.} \\
\midrule
Base & 0.818 & 0.781 & 0.836 & 0.718 & 0.491 & 0.729 \\
Tree only & 0.823 & 0.791 & 0.838 & \textbf{0.725} & 0.516 & 0.739 \\
Graph only & 0.879 & \textbf{0.807} & 0.843 & 0.710 & \textbf{0.547} & 0.757 \\
Tree+Graph & \textbf{0.906} & \textbf{0.807} & \textbf{0.845} & \textbf{0.725} & 0.519 & \textbf{0.760} \\
\bottomrule
\end{tabular*}
\caption{
Tree and graph evidence ablation.
All numbers are accuracy.
The table compares tree-only, graph-only, and tree-plus-graph evidence, and reports per-task accuracy with a macro average over the five evaluated tasks.
Bold denotes the best result in each column.
}
\label{tab:tree-graph-ablation}
\end{table*}

Table~\ref{tab:tree-graph-ablation} further shows that both structural views contribute useful information. Tree-only retrieval improves over the base model by providing hierarchical concept grounding, while graph-only retrieval gives larger average gains by exposing cross-branch clinical relations. Their combination achieves the best macro average, indicating that hierarchy and relation evidence are complementary.

\section{Additional Ablations}
\label{app:additional-ablation}
\paragraph{Structure update rule ablation.}
Table~\ref{tab:update-rule-ablation} compares three ways of combining initial structural confidence with LM-derived evidence: retaining the initial weights without refinement, simple linear interpolation, and the Beta-Bernoulli update used by TRACE. All variants use the same structure, retrieved evidence, and downstream inference procedure.

\begin{table}[t]
\centering
\footnotesize
\setlength{\tabcolsep}{4pt}
\renewcommand{\arraystretch}{1.05}
\begin{tabular}{lccc}
\toprule
Update rule
& \makecell{Exam\\QA}
& \makecell{Clinical\\records}
& \makecell{5-task\\Macro Avg.} \\
\midrule
Initial weight & 0.838 & \textbf{0.624} & \textbf{0.753} \\
Linear update & 0.841 & 0.615 & 0.751 \\
Beta-Bernoulli & \textbf{0.843} & 0.617 & 0.752 \\
\bottomrule
\end{tabular}
\caption{
Accuracy under different structure-weight update rules.
The results are similar across update mechanisms; Beta-Bernoulli is used for its bounded probabilistic interpretation and support for iterative refinement, rather than for a consistent accuracy advantage.
}
\label{tab:update-rule-ablation}
\end{table}

\paragraph{Backbone-level Tree-Relational RAG ablation.}
Table~\ref{tab:backbone-structure-ablation-appendix} applies the same learned structure to different backbone models through the same Tree-Relational RAG interface. The 10-task macro average improves for Baichuan-M2-32B, Baichuan-M3-235B, Llama-4-Scout, HuatuoGPT-3-32B, and Qwen3-32B. This cross-backbone improvement provides direct evidence that the learned tree-relational structure contains reusable oncology knowledge, rather than only fitting the Qwen3.5-27B backbone. DeepSeek-V4-Flash decreases on METABRIC, indicating that the benefit still depends on the backbone's calibration and sensitivity to prompt evidence.

\begin{table*}[!h]
\centering
\footnotesize
\setlength{\tabcolsep}{4.5pt}
\renewcommand{\arraystretch}{1.08}
\begin{tabular*}{\textwidth}{@{\extracolsep{\fill}}llcccc@{}}
\toprule
Model & Variant
& \makecell{Oncology\\examination QA}
& \makecell{Private clinical\\records}
& METABRIC
& \makecell{10-task\\Macro Avg.} \\
\midrule
\multirow{2}{*}{Baichuan-M2-32B}
& Base & 0.768 & 0.582 & 0.692 & 0.693 \\
& + Tree-Relational RAG & 0.779 {\scriptsize(+0.012)} & 0.578 {\scriptsize(-0.004)} & 0.705 {\scriptsize(+0.013)} & 0.702 {\scriptsize(+0.009)} \\
\midrule
\multirow{2}{*}{Baichuan-M3-235B}
& Base & 0.820 & 0.617 & 0.697 & 0.718 \\
& + Tree-Relational RAG & 0.845 {\scriptsize(+0.025)} & 0.601 {\scriptsize(-0.016)} & 0.705 {\scriptsize(+0.008)} & 0.726 {\scriptsize(+0.008)} \\
\midrule
\multirow{2}{*}{Llama-4-Scout}
& Base & 0.686 & 0.461 & 0.640 & 0.618 \\
& + Tree-Relational RAG & 0.785 {\scriptsize(+0.099)} & 0.559 {\scriptsize(+0.098)} & 0.647 {\scriptsize(+0.007)} & 0.671 {\scriptsize(+0.053)} \\
\midrule
\multirow{2}{*}{HuatuoGPT-3-32B}
& Base & 0.775 & 0.604 & 0.668 & 0.687 \\
& + Tree-Relational RAG & 0.820 {\scriptsize(+0.045)} & 0.603 {\scriptsize(-0.001)} & 0.687 {\scriptsize(+0.019)} & 0.710 {\scriptsize(+0.023)} \\
\midrule
\multirow{2}{*}{Qwen3-32B}
& Base & 0.783 & 0.610 & 0.673 & 0.694 \\
& + Tree-Relational RAG & 0.821 {\scriptsize(+0.038)} & 0.617 {\scriptsize(+0.007)} & 0.679 {\scriptsize(+0.006)} & 0.709 {\scriptsize(+0.015)} \\
\midrule
\multirow{2}{*}{DeepSeek-V4-Flash}
& Base & 0.836 & 0.615 & 0.688 & 0.718 \\
& + Tree-Relational RAG & 0.842 {\scriptsize(+0.007)} & 0.618 {\scriptsize(+0.003)} & 0.658 {\scriptsize(-0.031)} & 0.705 {\scriptsize(-0.013)} \\
\bottomrule
\end{tabular*}
\caption{
Backbone-level ablation of Tree-Relational RAG.
Each model is compared with and without the same learned structural retrieval module.
Numbers in parentheses indicate the change over the corresponding base model.
}
\label{tab:backbone-structure-ablation-appendix}
\end{table*}

\paragraph{Task-branch weighting ablation.}
We further ablate task-branch weighting in METABRIC mixed fine-tuning. Table~\ref{tab:task-branch-ablation} follows the same metric definitions as Table~\ref{tab:metabric-ft}. Removing task-branch weighting reduces overall accuracy from 0.7969 to 0.7614 and task-macro F1 from 0.5792 to 0.5376. This suggests that supervised task-branch weighting helps TRACE select task-appropriate structural evidence when multiple METABRIC tasks share the same learned oncology structure.

\begin{table*}[t]
\centering
\footnotesize
\setlength{\tabcolsep}{6pt}
\renewcommand{\arraystretch}{1.06}
\begin{tabular*}{\textwidth}{@{\extracolsep{\fill}}lcccc@{}}
\toprule
Method
& \makecell{Overall\\Acc.}
& \makecell{Overall\\F1}
& \makecell{Task-macro\\Acc.}
& \makecell{Task-macro\\F1} \\
\midrule
TRACE w/o task-branch weighting & 0.7614 & 0.3853 & 0.7628 & 0.5376 \\
TRACE & \textbf{0.7969} & \textbf{0.4802} & \textbf{0.7968} & \textbf{0.5792} \\
\bottomrule
\end{tabular*}
\caption{
Task-branch weighting ablation on METABRIC mixed fine-tuning.
Metrics follow the same definition as Table~\ref{tab:metabric-ft}.
}
\label{tab:task-branch-ablation}
\end{table*}

\section{Retrieval Baseline F1 Comparison}
\label{app:retrieval-baseline-f1}

Table~\ref{tab:retrieval-baseline-f1} reports the macro-F1 comparison between Vanilla RAG and Generic GraphRAG on the three examination QA tasks and two clinical-record tasks. Both retrieval baselines use the same oncology corpus as external evidence. Generic GraphRAG improves the task-macro F1 over Vanilla RAG, suggesting that graph-structured retrieval provides more useful evidence organization than passage retrieval alone.

\begin{table*}[t]
\centering
\footnotesize
\setlength{\tabcolsep}{4pt}
\renewcommand{\arraystretch}{1.05}
\begin{tabular*}{\textwidth}{@{\extracolsep{\fill}}lcccccc@{}}
\toprule
Method
& \makecell{Prof.\\Practice}
& \makecell{Oncology\\Attending}
& \makecell{Related\\Knowledge}
& \makecell{Common\\Cancer Dx.}
& \makecell{Breast\\Stage Pred.}
& \makecell{Task-macro} \\
\midrule
Vanilla RAG & 0.1380 & 0.7038 & 0.8624 & 0.4729 & \textbf{0.4546} & 0.5263 \\
Generic GraphRAG & 0.1365 & \textbf{0.7286} & \textbf{0.8815} & \textbf{0.4915} & 0.4495 & \textbf{0.5375} \\
\bottomrule
\end{tabular*}
\caption{
Macro-F1 comparison between retrieval baselines on the three examination QA tasks and two clinical-record tasks.
Both baselines use the same oncology corpus as external evidence.
Bold denotes the best result in each column.
}
\label{tab:retrieval-baseline-f1}
\end{table*}

\section{Leakage-Controlled METABRIC Analysis}
\label{app:metabric-leakage-control}
METABRIC is a retrospective structured-record benchmark, and non-target fields may include variables that are downstream of the prediction target, such as treatment or outcome information. To test whether TRACE mainly benefits from such reverse-causal shortcuts, we construct a stricter METABRIC variant by removing treatment-related and outcome-related fields from the serialized input, including endocrine therapy, chemotherapy, radiotherapy, surgery, recurrence, survival, follow-up, and death-related fields. The target label is withheld as in the main evaluation.

\begin{table*}[t]
\centering
\footnotesize
\setlength{\tabcolsep}{3.2pt}
\renewcommand{\arraystretch}{1.05}
\begin{tabular*}{\textwidth}{@{\extracolsep{\fill}}lcccccc}
\toprule
Method
& Stage
& Histology
& ER
& PR
& Relapse
& Avg. \\
\midrule
\multicolumn{7}{l}{\textit{Accuracy}} \\
Qwen3.5-27B & 0.662 & 0.762 & 0.755 & 0.744 & 0.594 & 0.704 \\
Vanilla RAG & 0.656 & 0.762 & 0.758 & \textbf{0.745} & 0.611 & 0.707 \\
Generic GraphRAG & 0.671 & 0.762 & 0.737 & 0.744 & 0.611 & 0.705 \\
TRACE & \textbf{0.741} & \textbf{0.763} & \textbf{0.777} & 0.744 & \textbf{0.620} & \textbf{0.729} \\
\midrule
\multicolumn{7}{l}{\textit{Macro-F1}} \\
Qwen3.5-27B & 0.356 & 0.108 & 0.715 & 0.722 & 0.583 & 0.497 \\
Vanilla RAG & 0.349 & 0.108 & 0.716 & 0.722 & 0.581 & 0.495 \\
Generic GraphRAG & 0.360 & 0.108 & 0.697 & 0.722 & 0.586 & 0.495 \\
TRACE & \textbf{0.405} & \textbf{0.192} & \textbf{0.728} & 0.722 & \textbf{0.591} & \textbf{0.527} \\
\bottomrule
\end{tabular*}
\caption{
Leakage-controlled METABRIC evaluation after removing treatment-related and outcome-related input fields.
All methods are evaluated on the same filtered inputs. Avg. reports task-macro average over the five METABRIC-derived tasks.
Bold denotes the best result in each column; ties after rounding are left unmarked.
}
\label{tab:metabric-no-treatment-outcome}
\end{table*}

Table~\ref{tab:metabric-no-treatment-outcome} shows that removing these fields lowers the absolute performance of several methods, confirming that treatment and outcome variables are informative in the original retrospective records. However, TRACE still achieves the best task-macro accuracy and macro-F1. It improves task-macro accuracy from 0.704 for Qwen3.5-27B and 0.707 for Vanilla RAG to 0.729, and improves task-macro F1 from 0.497 to 0.527. These results indicate that TRACE does not rely solely on treatment or outcome shortcuts; the learned tree-relational evidence remains useful under a stricter input setting.

\section{Additional Structural Evidence Case Studies}
\label{app:case-study}

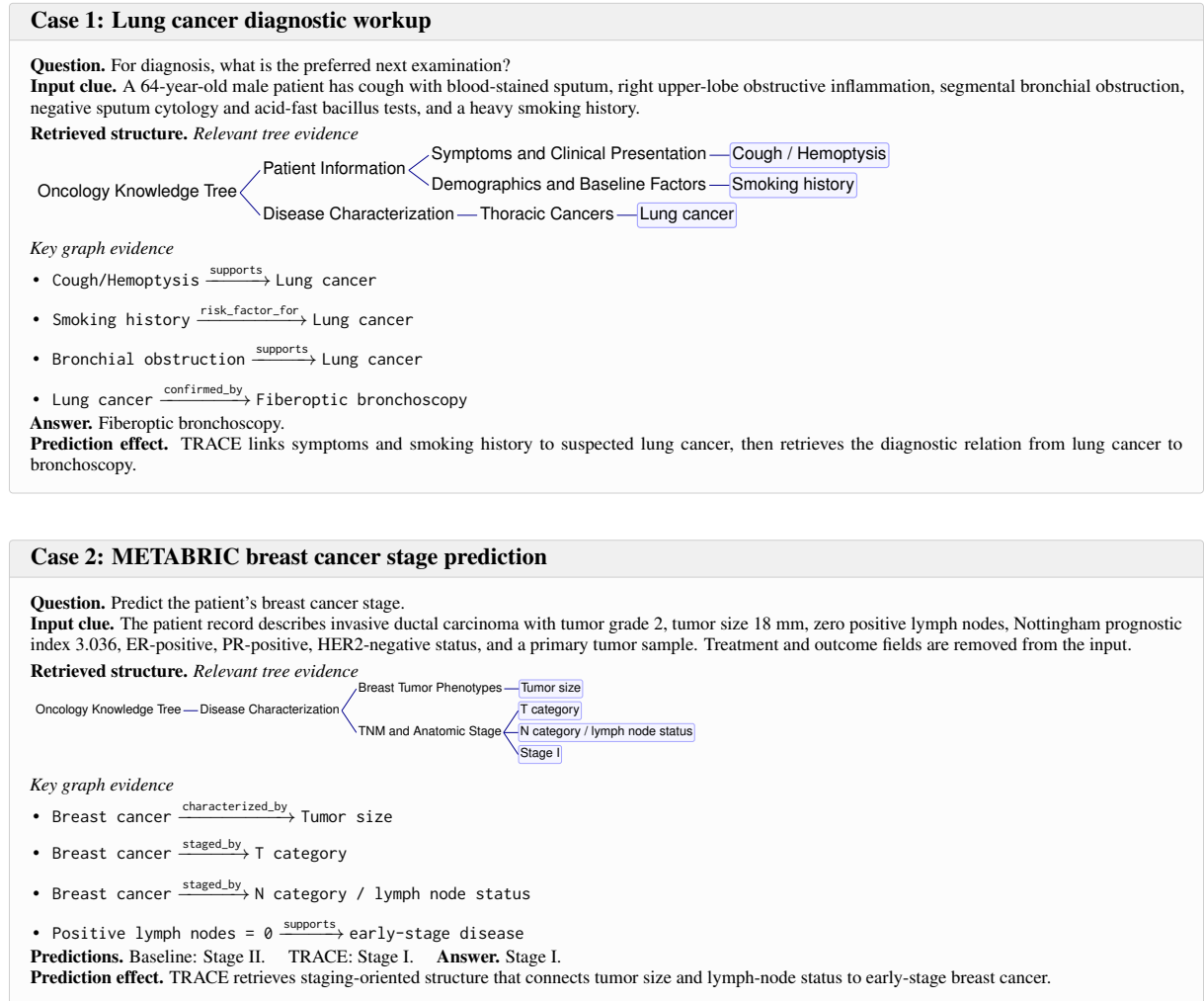
\begin{figure*}[!h]
\centering

\begin{tcolorbox}[tracecasewide,title={Case 1: Lung cancer diagnostic workup}]
\scriptsize
\textbf{Question.} For diagnosis, what is the preferred next examination?

\textbf{Input clue.} A 64-year-old male patient has cough with blood-stained sputum, right upper-lobe obstructive inflammation, segmental bronchial obstruction, negative sputum cytology and acid-fast bacillus tests, and a heavy smoking history.

\vspace{2pt}
\textbf{Retrieved structure.} \emph{Relevant tree evidence}

\resizebox{0.75\linewidth}{!}{
\begin{forest}
trace wide tree
[Oncology Knowledge Tree
  [Patient Information
    [Symptoms and Clinical Presentation
      [Cough / Hemoptysis]
    ]
    [Demographics and Baseline Factors
      [Smoking history]
    ]
  ]
  [Disease Characterization
    [Thoracic Cancers
      [Lung cancer]
    ]
  ]
]
\end{forest}
}

\vspace{2pt}
\emph{Key graph evidence}
\begin{itemize}[leftmargin=1.2em,itemsep=0pt,topsep=1pt]
    \item \texttt{Cough/Hemoptysis} \(\xrightarrow{\texttt{supports}}\) \texttt{Lung cancer}
    \item \texttt{Smoking history} \(\xrightarrow{\texttt{risk\_factor\_for}}\) \texttt{Lung cancer}
    \item \texttt{Bronchial obstruction} \(\xrightarrow{\texttt{supports}}\) \texttt{Lung cancer}
    \item \texttt{Lung cancer} \(\xrightarrow{\texttt{confirmed\_by}}\) \texttt{Fiberoptic bronchoscopy}
\end{itemize}

\textbf{Answer.} Fiberoptic bronchoscopy.

\textbf{Prediction effect.}
TRACE links symptoms and smoking history to suspected lung cancer, then retrieves the diagnostic relation from lung cancer to bronchoscopy.
\end{tcolorbox}

\vspace{4pt}

\begin{tcolorbox}[tracecasewide,title={Case 2: METABRIC breast cancer stage prediction}]
\scriptsize
\textbf{Question.} Predict the patient's breast cancer stage.

\textbf{Input clue.} The patient record describes invasive ductal carcinoma with tumor grade 2, tumor size 18 mm, zero positive lymph nodes, Nottingham prognostic index 3.036, ER-positive, PR-positive, HER2-negative status, and a primary tumor sample. Treatment and outcome fields are removed from the input.

\vspace{2pt}
\textbf{Retrieved structure.} \emph{Relevant tree evidence}

\resizebox{0.58\linewidth}{!}{
\begin{forest}
trace wide tree
[Oncology Knowledge Tree
  [Disease Characterization
    [Breast Tumor Phenotypes
      [Tumor size]
    ]
    [TNM and Anatomic Stage
      [T category]
      [N category / lymph node status]
      [Stage I]
    ]
  ]
]
\end{forest}
}

\vspace{2pt}
\emph{Key graph evidence}
\begin{itemize}[leftmargin=1.2em,itemsep=0pt,topsep=1pt]
    \item \texttt{Breast cancer} \(\xrightarrow{\texttt{characterized\_by}}\) \texttt{Tumor size}
    \item \texttt{Breast cancer} \(\xrightarrow{\texttt{staged\_by}}\) \texttt{T category}
    \item \texttt{Breast cancer} \(\xrightarrow{\texttt{staged\_by}}\) \texttt{N category / lymph node status}
    \item \texttt{Positive lymph nodes = 0} \(\xrightarrow{\texttt{supports}}\) \texttt{early-stage disease}
\end{itemize}

\textbf{Predictions.} Baseline: Stage II. \quad TRACE: Stage I.
\quad
\textbf{Answer.} Stage I.

\textbf{Prediction effect.}
TRACE retrieves staging-oriented structure that connects tumor size and lymph-node status to early-stage breast cancer. 
\end{tcolorbox}
\caption{
Structural evidence case studies.
TRACE retrieves compact task-relevant substructures, including local tree evidence and key graph relations, that connect input clues to the final prediction.
}
\label{fig:case-study-appendix}
\end{figure*}

Figure~\ref{fig:case-study-appendix} provides two representative examples of TRACE structural evidence. The examples show how TRACE combines local tree paths with key graph relations to support downstream predictions.

\end{document}